%% file: main.tex
\documentclass[runningheads]{llncs}

\usepackage[T1]{fontenc}
\usepackage{subcaption}
\usepackage{multirow}
\usepackage{hyperref}
\usepackage{tikz}
\usetikzlibrary{calc} 
\usepackage{orcidlink}
\usepackage[misc,geometry]{ifsym}

\DeclareCaptionLabelSeparator{custom}{\hspace{0.47mm}}
\usepackage{graphicx}
\begin{document}

\title{Self-supervised pretraining for cardiovascular magnetic resonance cine segmentation}

\titlerunning{Self-supervised pretraining for CMR cine segmentation}

\author{Rob A.J. de Mooij\textsuperscript{(\Letter)}\orcidlink{0009-0000-1014-3580} \and
Josien P.W. Pluim \and
Cian M. Scannell\orcidlink{0000-0001-9240-793X}}

\authorrunning{R.A.J. de Mooij et al.}

\institute{Department of Biomedical Engineering, Eindhoven University of Technology, Eindhoven, The Netherlands
\newline\email{r.a.j.d.mooij@tue.nl}\\
}

\maketitle             

\begin{abstract}
{\parindent15pt
\input{0_abstract}
}

\end{abstract}

\section{Introduction}
\input{1_introduction}

\section{Methods}
\input{2_methods}

\section{Results}
\input{3_results}

\section{Discussion}
\input{4_discussion}

\begin{credits}

\subsubsection{\discintname}
The authors have no competing interests to declare that are relevant to the content of this article. 

\end{credits}
\bibliographystyle{splncs04}
\bibliography{bibliography.bib}

\newpage
\appendix
{\Large
\begin{center}
    \textbf{Supplementary materials}
\end{center}}
\input{5_appendix}

\end{document}

%% file: 0_abstract.tex
Self-supervised pretraining (SSP) has shown promising results in learning from large unlabeled datasets and, thus, could be useful for automated cardiovascular magnetic resonance (CMR) short-axis cine segmentation. However, inconsistent reports of the benefits of SSP for segmentation have made it difficult to apply SSP to CMR. Therefore, this study aimed to evaluate SSP methods for CMR cine segmentation.

To this end, short-axis cine stacks of 296 subjects (90618 2D slices) were used for unlabeled pretraining with four SSP methods; SimCLR, positional contrastive learning, DINO, and masked image modeling (MIM). Subsets of varying numbers of subjects were used for supervised fine-tuning of 2D models for each SSP method, as well as to train a 2D baseline model from scratch. The fine-tuned models were compared to the baseline using the 3D Dice similarity coefficient (DSC) in a test dataset of 140 subjects.

The SSP methods showed no performance gains with the largest supervised fine-tuning subset compared to the baseline (DSC = 0.89). When only 10 subjects (231 2D slices) are available for supervised training, SSP using MIM (DSC = 0.86) improves over training from scratch (DSC = 0.82).

This study found that SSP is valuable for CMR cine segmentation when labeled training data is scarce, but does not aid state-of-the-art deep learning methods when ample labeled data is available. Moreover, the choice of SSP method is important. The code is publicly available at: \url{https://github.com/q-cardIA/ssp-cmr-cine-segmentation}

%% file: 1_introduction.tex
Self-supervised learning allows deep learning models to learn useful information from unlabeled data \cite{balestriero_cookbook_2023}. Self-supervised and traditional supervised learning are often combined for specific tasks, where a model first trains on unlabeled data with self-supervised pretraining (SSP), before supervised fine-tuning on labeled data for a specific downstream task \cite{zoph_rethinking_2020}. SSP could be especially important in medical imaging, due to the lack of high quality labeled datasets.

Cardiovascular magnetic resonance (CMR) short-axis cine segmentation is an interesting problem on which to investigate SSP, both from a methodological and clinical perspective. CMR short-axis cine data consists of 4D (3D + time) images, allowing SSP methods that incorporate spatial and temporal information into the pretraining task \cite{kalapos_self-supervised_2022,zeng_positional_2021}. Additionally, labeled short-axis cine datasets consist of many unlabeled images as only certain cardiac phases are annotated. Therefore, models can be pretrained with SSP on all available images and fine-tuned on the labeled cardiac phases. Finally, improving the accuracy of CMR segmentation, potentially with SSP, could improve the evaluation of cardiac structure and function in the clinic \cite{davies_precision_2022,sirajuddin_ischemic_2021}.

However, SSP has yielded conflicting results in its application in medical imaging. Many SSP methods have been proposed for medical imaging, but many results have been difficult to generalize to different problems and data \cite{zhang_dive_2023}. Medical classification problems have seen multiple promising SSP methods \cite{huang_self-supervised_2023}. However, benefits of SSP in segmentation tasks often hold only with very limited labeled training data \cite{kalapos_self-supervised_2022,vanberlo_survey_2024,zhang_dive_2023}. Additionally, it is difficult to disentangle the effect of SSP from that of novel architectures such as vision transformers (ViT), and results depend on data and training hyperparameters \cite{zhang_dive_2023}. These inconsistent findings make it challenging to choose the most effective SSP method in medical imaging problems, especially for segmentation tasks.

This study investigates SSP methods for CMR cine segmentation with convolutional neural networks (CNNs), and the effectiveness of SSP methods with varying amounts of labeled data available during supervised fine-tuning. In particular, we will: (1) develop a strong CMR cine segmentation model to serve as baseline, (2) optimize four promising SSP methods for unlabeled pretraining of the baseline model, (3) compare the fine-tuned models, after SSP, to the baseline in downstream segmentation performance with varying amounts of labeled fine-tuning data, and (4) investigate the effects of data augmentation and SSP on model generalizability.

%% file: 2_methods.tex
\subsection{Data}
Short-axis cine data of the M\&Ms \cite{campello_multi-centre_2021} and M\&Ms-2 \cite{martin-isla_deep_2023} challenges were used for a total of 510 subjects. For each subject, a 4D cine image was available with a varying number of slices (median 11) and time frames (median 25). Each cine image had labels for the end diastole (ED) and end systole (ES) time frames, delineating the left ventricle (LV), myocardium (MYO), and right ventricle (RV). All 345 publicly available subjects of the M\&Ms challenge data were included. Only 165 subjects of the M\&Ms-2 challenge data were included to avoid possible overlapping subjects.

Data were split on a subject level into 296 training, 74 validation, and 140 test subjects. First, 60 test subjects were selected from the M\&Ms data, using random sampling stratified by MR scanner vendor. The 80 test subjects for the M\&Ms-2 challenge data were obtained from the original challenge test set, after removing the subjects with possible overlap. The remaining 370 subjects were then randomly divided into training and validation using an 80/20\% split.

The training set consisted of 90618 2D slices, of which 6738 were labeled. To simulate the effect of smaller labeled datasets, three subsets of 50, 25, 15 and 10 random training subjects were used, consisting of on average 1151, 582, 349, and 231 slices respectively, depending on the subject subset. The labeled validation dataset was used for hyperparameter optimization, while the labeled test dataset was only used for the final evaluation.

\subsection{Baseline model}
A 2D fully convolutional U-Net baseline model was developed based on the nnU-Net method \cite{isensee_nnu-net_2021}. All hyperparameters were set based on the automatic configuration and CMR data experiments of the original nnU-Net work, including deep supervision, with more extreme data augmentation implemented to improve robustness and stability. The baseline was trained from scratch separately on the fully labeled training dataset and its subsets, which was repeated with three training seeds. This model setup served as a realistic baseline with a competitive performance in 2D cine segmentation against which to compare the SSP methods.

\subsection{Self-supervised pretraining methods}
\begin{figure*}[h]
    \begin{tikzpicture}[overlay]
        \draw[dashed] ($(0.3254\linewidth, 6.59)$) -- ($(0.3254\linewidth, 0.59)$);
        \draw[dashed] ($(0.6487\linewidth, 6.59)$) -- ($(0.6487\linewidth, 0.59)$);
        \draw[dashed] ($(0.8910\linewidth, 6.59)$) -- ($(0.8910\linewidth, 0.59)$);
    \end{tikzpicture}
    \begin{subfigure}{0.3097826087\linewidth}
        \centering
        \includegraphics[width=\linewidth]{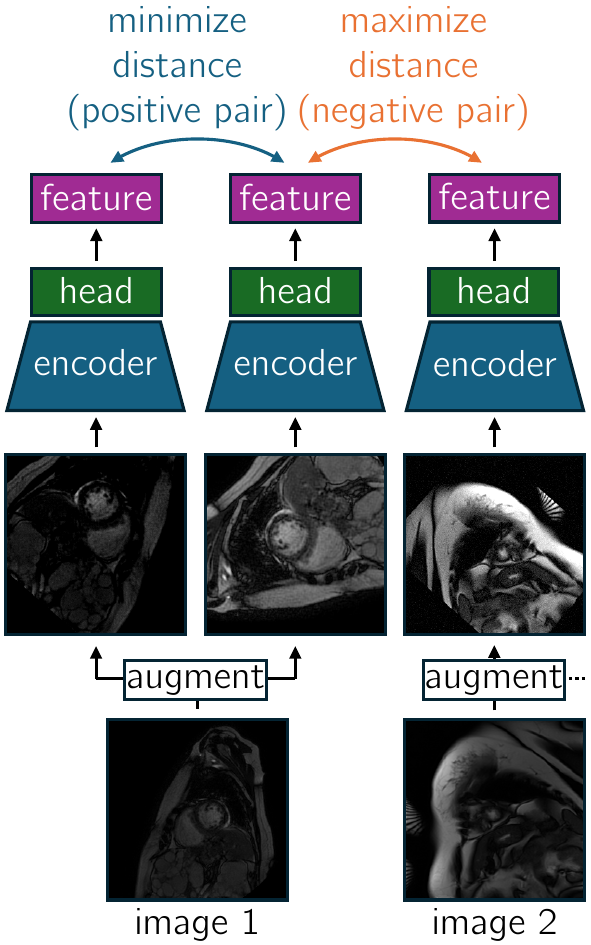}
        \captionsetup{justification=centering}
        \caption{SimCLR}        
        \label{fig:simclr}
    \end{subfigure}%
    \hfill
    \begin{subfigure}{0.3097826087\linewidth}
        \centering
        \includegraphics[width=\linewidth]{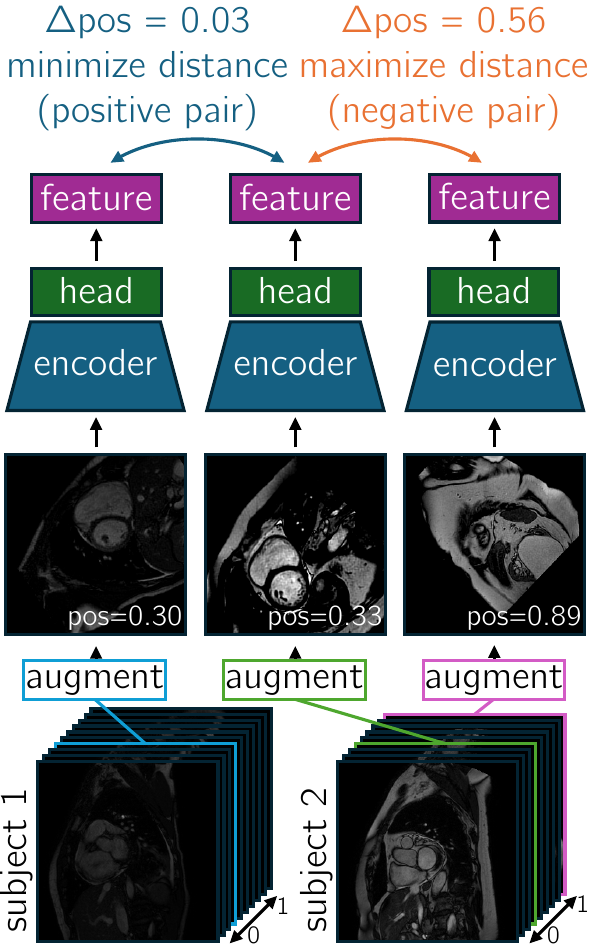}
        \captionsetup{justification=centering}
        \caption{PCL}
        \label{fig:pcl}
    \end{subfigure}%
    \hfill
    \begin{subfigure}{0.227173913\linewidth}
        \centering
        \includegraphics[width=\linewidth]{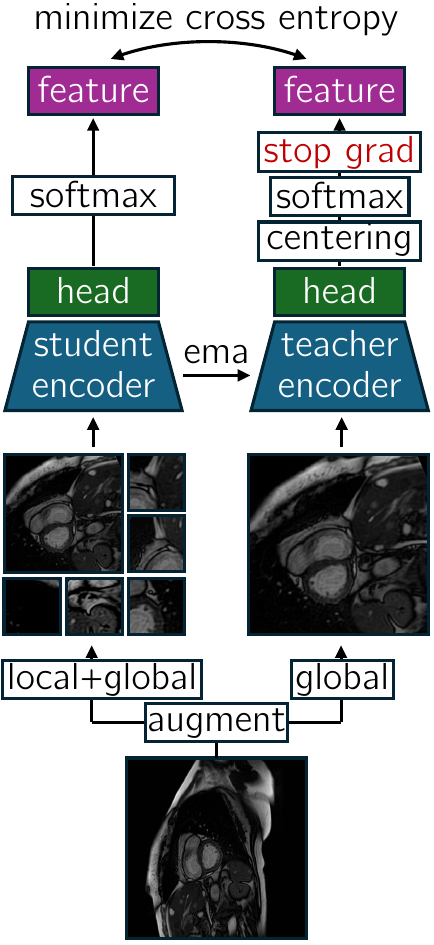}
        \captionsetup{justification=centering}
        \caption{DINO}
        \label{fig:dino}
    \end{subfigure}%
    \hfill
    \begin{subfigure}{0.1032608696\linewidth}
        \centering
        \includegraphics[width=\linewidth]{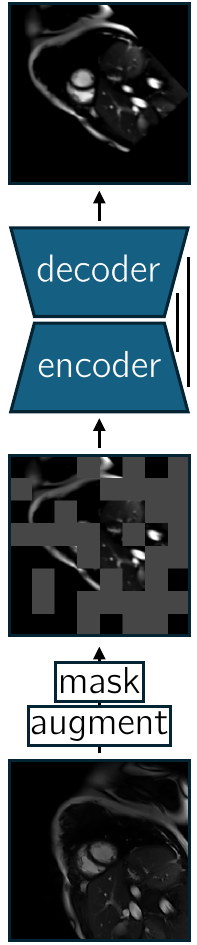}
        \captionsetup{justification=centering,labelsep=custom}
        \caption{MIM}
        \label{fig:mim}
    \end{subfigure}
\caption{Visualizations of the four SSP methods that were used.}
\label{fig:ssp_methods}    
\end{figure*}
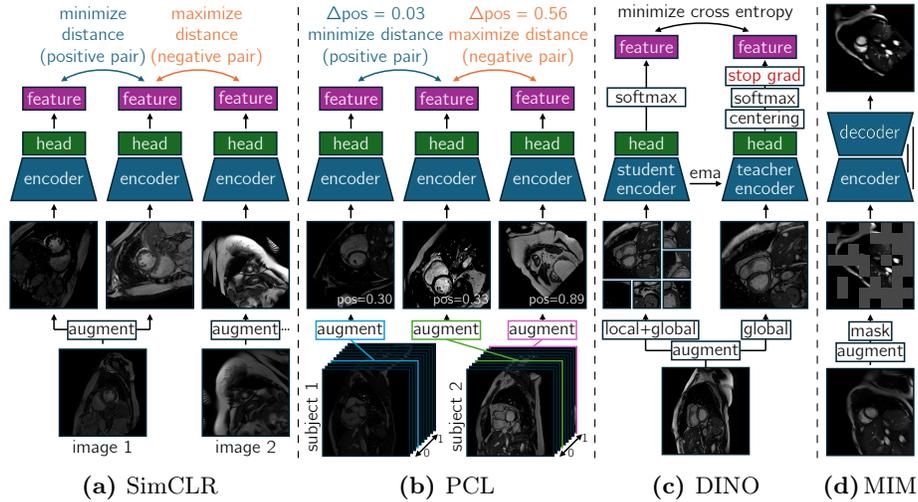

Four SSP methods were compared; \textit{a simple framework for contrastive learning of visual representations} (SimCLR) \cite{chen_simple_2020}, \textit{positional contrastive learning} (PCL) \cite{zeng_positional_2021}, \textit{self-\textbf{di}stillation with \textbf{no} labels} (DINO) \cite{caron_emerging_2021}, and \textit{masked image modeling} (MIM) \cite{chen_masked_2023,dominic_improving_2023,he_masked_2021}. These methods are illustrated in figure \ref{fig:ssp_methods}. SimCLR, PCL, and DINO were used to pretrain the encoder of the 2D U-Net, while MIM was used to pretrain both the encoder and decoder of the 2D U-Net.

Contrastive learning methods like SimCLR and PCL aim to train the model to extract useful features from the images by contrasting positive and negative image pairs. It aims for a feature space where positive image pairs are close together, while negative image pairs are further apart \cite{zhang_dive_2023}. For SimCLR and PCL, positive and negative image pairs are obtained from a training batch of augmented images, containing two instances of the same image with different data augmentations. For SimCLR, only the two augmented versions of the same image are considered a positive pair, while every other image pair is a negative pair \cite{chen_simple_2020}. In PCL, each 2D image is assigned a relative position in the range [0, 1], indicating its relative position in the 3D volume it originated from. An image pair is considered positive when the difference between their relative positions is below a certain threshold \cite{zeng_positional_2021}. 

With DINO, a student network learns from a teacher network how to predict global image features from local image patches \cite{caron_emerging_2021}. For each image, eight augmented copies are used; two global crops, and six local crops. The teacher network receives the two global crops, while the student network receives both the global and local crops. The student is trained to output the same as the teacher, while the teacher is only updated with the exponential moving average of the student network. 

MIM trains a model to restore an image that has been partially masked. An augmented image is first divided into equally sized patches, after which some of these patches are masked out. Patches are masked out by setting their pixel intensities to a predetermined value \cite{dominic_improving_2023}.

\subsection{Pretraining}
 All methods were pretrained, without labels, on the full training dataset of 90618 slices. Pretraining hyperparameters were based on the hyperparameters from the original implementation. Hyperparameters were further fine-tuned for each method based on pretraining stability and cine segmentation performance after fine-tuning. Choices that worked well for all pretraining methods included: (1) 100 epochs, (2) stochastic gradient descent (SGD) optimizer, (3) extensive data augmentation, and (4) cosine learning rate scheduler. Details for all model trainings can be found in supplementary material \ref{appendix:hyperparameters}.
 
 SimCLR had no method-specific hyperparameters, while the relative position threshold for positive pairs in PCL was chosen as 0.1. For DINO, the original hyperparameters from the CNN experiments were used for reference \cite{caron_emerging_2021}. Global and local patch sizes were increased to 256x256 and 128x128 respectively due to the deep CNN architecture requiring the image size to be divisible by 64. Finally, the DINO head output dimensionality was reduced to 8192.
 
 The MIM hyperparameters were adapted based on multiple similar implementations \cite{chen_masked_2023,dominic_improving_2023,he_masked_2021}. A patch size of 32x32 was used, with a mask ratio of 0.75, and a constant masking value of 0.0, applied after image standardization and data augmentation. The architecture was the same as that of the 2D baseline model, except for the removal of the additional outputs, as deep supervision was not used during MIM pretraining. Transferring the weights of the encoder and decoder, but without the output convolution at the end, resulted in the best downstream performance. A mean squared error loss was used, applied only on the masked patches.

\subsection{Fine-tuning}
After SSP with all available images, models were fine-tuned separately on all labeled time frames and subsets of that. Fine-tuning was repeated three times for each pretrained model, changing the training seed that determined data loading, data augmentation, and model weight initialization for weights that had not been pretrained. The training seed was also used to select the subjects in the subset experiments. Fine-tuning used the hyperparameters of the fully-supervised training of the baseline from scratch with the learning rate lowered to 0.005.

\subsection{Generalization evaluation}
The best performing SSP method was further investigated to give insights into its behaviour. Two experiments investigated the generalizability of fine-tuned models after SSP on all data. For each experiment, the best performing SSP method was used to pretrain on all available data, while fine-tuning and test data were varied. Data augmentation was investigated separately, for a total of three experiments:

\textbf{Generalization to unseen cardiac phases.} To evaluate the benefit of SSP in generalizing to data unseen during fine-tuning, we limited the fine-tuning data to a single cardiac phase, allowing for out-of-domain evaluation. A baseline model was trained from scratch on a single time frame per subject for reference. Then, the model pretrained on all data was fine-tuned on a single time frame per subject. Each experiment was performed separately for ED and ES time frames. 

\textbf{Generalization to unseen vendors.} A similar experiment was performed to investigate the generalizability to vendors unseen during fine-tuning. The four vendors of the M\&Ms challenge datasets were divided into two groups: Siemens and Philips (A+B), and General Electric and Canon (C+D). Baseline training and SSP fine-tuning were trained on both vendor groups separately. The in-domain and out-of-domain performance of the resulting models was compared separately on the test data.

\textbf{Generalization from data augmentation.} So far, all models were trained with data augmentation, making it difficult to know to what extent model performance and robustness was due to SSP, and to what extent to data augmentation. Therefore, this final experiment trained versions of the baseline and best performing SSP method without data augmentation during training from scratch and fine-tuning respectively.

\subsection{Evaluation}
Segmentation performance was evaluated with the Dice similarity coefficient (DSC) in 3D, for both the ED and ES time frames. The metric was calculated separately for the LV, MYO and RV classes. For each fine-tuned model, the mean DSC across the three classes and for all test subjects was calculated. The SSP methods were compared with the baseline model for the three random training seeds. For the generalization experiments, test subsets for different cardiac phases and vendors were also evaluated separately.

All pretraining and fine-tuning used custom implementations that are available at \url{https://github.com/q-cardIA/ssp-cmr-cine-segmentation}.

%% file: 3_results.tex
The mean test DSCs across all foreground classes and varying numbers of fine-tuning subjects can be seen in table \ref{tab:subset_dice}. All models show similar performances when trained or fine-tuned on all available labeled training data. This also holds when looking at individual classes. All models showed mean 3D test DSCs of 0.93, 0.85, and 0.90 for the LV, MYO, and RV classes respectively. More details are shown in supplementary material \ref{appendix:class_dice}.

\begin{table}[h]
\caption{Mean test DSC across all 140 test subjects and foreground classes for the baseline and fine-tuned SSP models, for varying numbers of labeled subjects ± sample standard deviation across different training seeds. For each number of labeled subjects, the average number of slices across all training seeds is shown. The biggest performance increase compared to the baseline is indicated in bold.}%
\vspace{4pt}
\centering
\begin{tabular*}{\linewidth}{@{\extracolsep{\fill}} lccccc }
\textbf{\begin{tabular}[c]{@{}l@{}}\#subjects\\ (\#slices)\end{tabular}} & \textbf{Baseline} & \textbf{SimCLR} & \textbf{PCL} & \textbf{DINO} & \textbf{MIM} \\
296 (6738)                                                               & 0.89 ± 0.001      & 0.89 ± 0.002    & 0.89 ± 0.003 & 0.89 ± 0.001  & 0.89 ± 0.001 \\
50 (1151)                                                                & 0.87 ± 0.007      & 0.87 ± 0.011    & 0.88 ± 0.005 & 0.87 ± 0.006  & 0.88 ± 0.006 \\
25 (582)                                                                 & 0.86 ± 0.003      & 0.86 ± 0.014    & 0.85 ± 0.014 & 0.84 ± 0.002  & 0.87 ± 0.007 \\
15 (349)                                                                 & 0.84 ± 0.007      & 0.83 ± 0.024    & 0.85 ± 0.009 & 0.79 ± 0.009  & 0.86 ± 0.015 \\
10 (231)                                                                 & 0.82 ± 0.009      & 0.80 ± 0.024    & 0.84 ± 0.009 & 0.74 ± 0.022  & \textbf{0.86 ± 0.007}
\end{tabular*}
\label{tab:subset_dice}
\end{table}

As shown in table \ref{tab:subset_dice}, PCL and MIM pretraining outperform the baseline performance for smaller labeled fine-tuning subsets. The biggest increase compared to the baseline can be seen for MIM pretraining and fine-tuning on the smallest labeled subset. SimCLR and DINO pretraining, on the other hand, show a performance decrease for fine-tuning on smaller labeled subsets, as well as a higher sample standard deviation.

Since MIM pretraining consistently showed the best results, MIM was used for the generalization evaluation. Tables \ref{tab:time_frames_dice} and \ref{tab:vendor_dice} show the results for the first two generalization experiments, investigating fine-tuned model generalization to unseen (during fine-tuning) cardiac phases and vendors. Both experiments showed similar overall, in-domain, and out-of-domain baseline performances. For labeled ED time frames fine-tuning, both overall and out-of-domain performances increased slightly with SSP. Contrastingly, labeled ES time frames fine-tuning only showed a slight in-domain performance increase with SSP. SSP did not result in changes in performance when fine-tuning on labeled vendors A and B. A slight increase in out-of-domain performance can be seen after fine-tuning on labeled vendors C and D.

\begin{table}[h]
\caption{Mean test DSC ± sample standard deviation, comparing fine-tuned model generalization to unseen cardiac phases. Improvements after SSP are indicated in bold.}%
\vspace{4pt}
\centering
\begin{tabular*}{\linewidth}{@{\extracolsep{\fill}} llccc }
\textbf{Pretraining} & \textbf{Fine-tuning} & \textbf{\begin{tabular}[c]{@{}c@{}}Both time \\ frames\end{tabular}} & \textbf{\begin{tabular}[c]{@{}c@{}}ED time \\ frames\end{tabular}} & \textbf{\begin{tabular}[c]{@{}c@{}}ES time \\ frames\end{tabular}} \\
None                    & ED time frames      & 0.88 ± 0.001       & 0.90 ± 0.002      & 0.86 ± 0.002      \\
All data             & ED time frames      & \textbf{0.89 ± 0.001}      & 0.90 ± 0.001      & \textbf{0.87 ± 0.001}     \\
None                    & ES time frames      & 0.88 ± 0.007       & 0.88 ± 0.012      & 0.88 ± 0.002      \\
All data             & ES time frames      & 0.88 ± 0.001       & 0.88 ± 0.001      & \textbf{0.89 ± 0.000}
\end{tabular*}
\label{tab:time_frames_dice}
\end{table}

\begin{table}[h]
\caption{Mean test DSC ± sample standard deviation, comparing fine-tuned model generalization to unseen vendors. Improvements after SSP are indicated in bold.}%
\vspace{4pt}
\centering
\begin{tabular*}{\linewidth}{@{\extracolsep{\fill}} llccc }
\textbf{Pretraining} & \textbf{Fine-tuning} & \textbf{All vendors} & \textbf{Vendors A+B}  & \textbf{Vendors C+D} \\
None                    & Vendors A+B         & 0.89 ± 0.001           & 0.89 ± 0.001          & 0.88 ± 0.002         \\
All data             & Vendors A+B         & 0.89 ± 0.001           & 0.89 ± 0.001          & 0.88 ± 0.000         \\
None                    & Vendors C+D         & 0.88 ± 0.002           & 0.87 ± 0.001          & 0.89 ± 0.004         \\
All data             & Vendors C+D         & 0.88 ± 0.001           & \textbf{0.88 ± 0.002} & 0.89 ± 0.002         
\end{tabular*}
\label{tab:vendor_dice}
\end{table}

The baseline mean test DSC (± sample standard deviation across training seeds) of 0.89 ± 0.001 decreased to 0.69 ± 0.039 when trained without data augmentation. With MIM pretraining, the mean test DSC of 0.89 ± 0.001 lowered to 0.78 ± 0.040 without data augmentation, a smaller decrease than for the baseline model.

The DSCs for varying cardiac phase and vendor subsets for baseline models trained on all labeled data with and without data augmentation can be seen in table \ref{tab:data_augmentation_dice}. This shows that without data augmentation, there is a large difference in DSC between ED and ES time frames, as well as between both vendor groups. With data augmentation the performance gap between times frames is mostly bridged, while the performance gap between vendors is bridged completely.

\begin{table}[h]
\caption{Mean test DSC ± standard deviation, for varying cardiac phases and vendor subsets, for baseline models with and without data augmentation.}%
\vspace{4pt}
\begin{tabular*}{\linewidth}{@{\extracolsep{\fill}} lcccc }
\textbf{Data augmentation} & \textbf{\begin{tabular}[c]{@{}c@{}}ED time\\ frames\end{tabular}} & \textbf{\begin{tabular}[c]{@{}c@{}}ES time\\ frames\end{tabular}} & \textbf{\begin{tabular}[c]{@{}c@{}}Vendors\\ A+B\end{tabular}} & \textbf{\begin{tabular}[c]{@{}c@{}}Vendors\\ C+D\end{tabular}} \\
Yes     & 0.90 ± 0.001       & 0.88 ± 0.003      & 0.89 ± 0.001      & 0.89 ± 0.006      \\
No      & 0.71 ± 0.042       & 0.66 ± 0.036      & 0.72 ± 0.036      & 0.61 ± 0.049
\end{tabular*}
\label{tab:data_augmentation_dice}
\end{table}

%% file: 4_discussion.tex
This study aimed to investigate SSP methods for CMR cine segmentation with CNNs, and the effectiveness of SSP methods with varying amounts of fine-tuning data. More specifically, four SSP methods were compared to a baseline for varying numbers of labeled subjects for fine-tuning. Additionally, further experiments investigated generalizability of fine-tuned models for unseen cardiac phases and MR scanner vendors after SSP, and generalizability due to data augmentation.

Table \ref{tab:subset_dice} indicates that SSP with unlabeled data only yields an improvement in CMR cine segmentation when very limited amounts of labeled data are available for fine-tuning. Moreover, choice of SSP method is important. PCL and MIM showed performance increases for small labeled fine-tuning datasets, indicating that these methods contribute useful information for the problem. MIM may be the most effective because it pretrains both the encoder and decoder of the U-Net architecture.

SimCLR and DINO show decreased performances for limited amounts of labeled data, with a generally higher standard deviation. DINO pretraining shows the largest performance decrease, a potential explanation is that it is ill-suited for CNN architectures, as it is designed for ViTs. Additionally, both SimCLR and DINO generally require large batch sizes during training which was limited by our available hardware \cite{caron_emerging_2021,chen_simple_2020}.

The cardiac phase generalization experiment shows a slight benefit in SSP in generalizing to unseen (during fine-tuning) cardiac phases. Out-of-domain performances can slightly increase with SSP, while in-domain performances did not decrease. This indicates that there may be benefit in SSP in this situation. However, the baseline models already show the ability to generalize to unseen cardiac phases, leaving little room for improvement as a result of SSP. This could be explained by the data augmentation. Table \ref{tab:data_augmentation_dice} shows that there is a performance gap between cardiac phases when training on all labeled data without data augmentation. However, adding data augmentation largely closes this gap, indicating that data augmentation accounts for most of the generalizability shown in the unseen cardiac phase experiment. 

Similar results can be seen for the unseen vendor experiment, showing a small out-of-domain performance increase. For the two vendor groups, table \ref{tab:data_augmentation_dice} also shows a large performance gap without data augmentation, which is closed with data augmentation. While these models were trained on all labeled data, these results do indicate the importance of data augmentation in general model performance and generalizability to unseen data. The results of the data augmentation experiment in combination with MIM pretraining further show the importance of data augmentation, even when using SSP. While SSP shows a clear benefit when not using data augmentation in fine-tuning and training from scratch, data augmentation is still necessary to achieve the best performance. This indicates that data augmentation can better cover the data distribution compared to our SSP methods with larger unlabeled datasets. These results support claims that data augmentation can mostly meet or exceed the benefits of SSP, when appropriately selected for the downstream task \cite{zhang_dive_2023}. However, our results also indicate that data augmentation is a crucial step in enabling the possible benefits of SSP.

Future research should further investigate whether these findings hold when using other model architectures, including ViTs. SSP methods such as DINO and MIM are generally developed for and perform better on ViTs \cite{caron_emerging_2021,chen_masked_2023}. However, this study intentionally focused on CNN architectures, to separate the effects of SSP from new model architectures.

In conclusion, SSP can be beneficial for CMR cine segmentation with limited amounts of labeled data, but its effectiveness depends on the SSP method. Additionally, SSP with large unlabeled datasets can provide slight benefits in generalizing to unseen domains for which labeled data is not available but unlabeled data is available for SSP.

%% file: 5_appendix.tex
\section{Training details}
\label{appendix:hyperparameters}
Hyperparameters for each model training. The DINO learning rate is reported after applying the linear scaling rule based on batch size.
\vspace{-4mm}
\begin{table}[]
\begin{tabular*}{\linewidth}{@{\extracolsep{\fill}} lcccccc }
\textbf{Hyperparameter}   & \textbf{Baseline} & \textbf{SimCLR} & \textbf{PCL} & \textbf{DINO} & \textbf{MIM} & \textbf{Fine-tuning} \\
Epochs  & 1000  & 100  & 100  & 100  & 100  & 1000  \\
Learning rate             & 0.01              & 0.1             & 0.1          & 0.0075       & 0.01         & 0.005               \\
Scheduler  & Polynomial  & Cosine  & Cosine  & Cosine  & Cosine  & Polynomial  \\
Optimizer  & SGD  & SGD  & SGD  & SGD  & SGD  & SGD  \\
Nesterov momentum     & yes               & no              & no           & no            & yes          & yes                 \\
Momentum              & 0.99              & 0.9             & 0.9          & 0.9           & 0.99         & 0.99                \\
Weight decay              & 3.0e-05           & 1.0e-04         & 1.0e-05      & 1.0e-04       & 3.0e-05      & 3.0e-05             \\
Batch size                & 32                & 224             & 64           & 64            & 128          & 32                  \\
Mixed precision & no                & yes             & no           & yes           & yes          & no                 
\end{tabular*}
\end{table}

\vspace{-4mm}

\section{Test DSC per class}
\label{appendix:class_dice}
Mean test DSC for the baseline and fine-tuned SSP models, for separate foreground classes ± mean sample standard deviation across all 140 test subjects.
\vspace{-8mm}
\begin{table}[h]
\begin{tabular*}{\linewidth}{@{\extracolsep{\fill}} lccccc }
\multicolumn{1}{c}{\textbf{Class}} & \textbf{Baseline} & \textbf{SimCLR} & \textbf{PCL} & \textbf{DINO} & \textbf{MIM} \\
LV                                 & 0.93 ± 0.047      & 0.93 ± 0.047    & 0.93 ± 0.048 & 0.93 ± 0.048  & 0.93 ± 0.048 \\
MYO                                & 0.85 ± 0.055      & 0.85 ± 0.048    & 0.85 ± 0.052 & 0.85 ± 0.052  & 0.85 ± 0.053 \\
RV                                 & 0.90 ± 0.062      & 0.90 ± 0.064    & 0.90 ± 0.061 & 0.90 ± 0.068  & 0.90 ± 0.068
\end{tabular*}
\end{table}